\documentclass{article}

\usepackage{PRIMEarxiv}

\usepackage[utf8]{inputenc}
\usepackage[T1]{fontenc}
\usepackage{amsmath,amssymb,amsfonts}
\usepackage{booktabs}
\usepackage{makecell}
\usepackage{graphicx}
\usepackage{subcaption}
\usepackage{microtype}
\usepackage{url}
\usepackage{natbib}
\usepackage{hyperref}
\usepackage{fancyhdr}
\usepackage{float}

\graphicspath{{Images/}}

\title{Trust-Aware Predictive Emissions Monitoring for Gas Turbine Fleets with Limited Labelled Data}

\author{
Rebecca Potts\\
School of Natural and Computing Sciences\\
University of Aberdeen, Aberdeen, UK\\
\texttt{r.potts.21@abdn.ac.uk}
\And
Aiden Durrant\\
School of Computing Sciences\\
University of East Anglia, UK\\
\texttt{Aiden.Durrant@uea.ac.uk}
\And
Rick Hackney\\
Siemens Energy Industrial Turbomachinery Ltd.\\
Lincoln, UK\\
\texttt{richard.hackney@siemens-energy.com}
\And
Georgios Leontidis\\
Department of Physics and Technology\\
UiT The Arctic University of Norway, Norway\\
\texttt{georgios.leontidis@uit.no}
}

\begin{document}
\maketitle

\begin{abstract}
Machine learning-based predictive emissions monitoring systems offer a practical alternative to direct emissions measurement, but their deployment across gas turbine fleets is challenging when emissions labels are available for only a small subset of assets. In this work, a trust-aware probabilistic framework is proposed for fleet-level gas turbine NOx prediction under limited labelled supervision. The framework combines a multi-head recurrent prediction model with learned confidence estimation, ensemble-based uncertainty quantification, auxiliary feature prediction, feature-space distance analysis, and operating-range diagnostics. These signals are calibrated on labelled data to produce interpretable per-sample trust scores, providing indicators of prediction reliability on unlabelled turbines, supporting the identification of predictions that should be treated with greater caution during fleet-level deployment. Confidence-based filtering reduces MAE from 0.202 at full coverage to 0.070 for the highest-confidence 10\% of predictions, demonstrating that confidence estimates are meaningfully related to prediction error. Unlabelled and out-of-distribution samples exhibit increased uncertainty and reduced confidence, indicating that the framework responds appropriately to distributional shift. The results show that the proposed trust framework provides actionable reliability information for emissions prediction on unlabelled turbines, supporting more transparent and trustworthy deployment of PEMS across industrial fleets.
\end{abstract}


\section{Introduction}\label{sec:intro}

Predictive emissions monitoring systems (PEMS) provide an alternative to direct emissions measurement, continuous emissions monitoring systems (CEMS), by estimating pollutant concentrations from available operation data. Traditional approaches may incorporate physical knowledge of the combustion process and turbine behaviour (\cite{hackney2016predictive}), while data-driven methods use machine learning models to learn relationships between operating conditions and emissions outputs (\cite{kaya2019predicting}, \cite{potts2023tabular}). These approaches are particularly valuable in industrial settings where continuous emissions measurements may be costly or unavailable across the fleet.

Machine learning-based PEMS requires ground-truth emissions measurements for training and validation. This presents a significant challenge in fleet-level deployment, where emissions labels may only be available for a small subset of turbines, such that reliable performance is not guaranteed on other gas turbines operating under different conditions. As a result, prediction accuracy on the labelled test set alone is insufficient: the reliability of each prediction must also be assessed. 

In this work, we consider a fleet-level emissions prediction problem in which labelled emissions data are available for a single turbine only, while 56 additional turbines contain operational data without associated emissions measurements. The aim is not only to predict emissions on these unlabelled turbines, but also to provide an interpretable assessment of prediction trustworthiness. This enables predictions to be used with greater confidence when the model is operating in familiar conditions, while highlighting cases where predictions should be treated with caution. 

We propose a trust-aware PEMS framework that learns a calibrated relationship between prediction error and model-derived reliability signals on labelled data. These signals include predictive uncertainty, epistemic uncertainty, learned confidence, feature-prediction behaviour, and distance from the labelled training distribution. By combining these signals into a calibrated trust score, the framework enables per-sample confidence estimation and reliability assessment under unlabelled operating conditions.

\section{Related Work}

\subsection{Machine Learning for Predictive Emissions Monitoring}

Predictive emissions monitoring systems (PEMS) are commonly used as validation or backup systems for continuous emissions monitoring systems (CEMS), particularly where direct measurement is costly or limited, offering greater flexibility than first-principles approaches when modelling complex and non-linear relationships between operating conditions and emissions. 

A number of machine learning methods have been used to predict emissions for gas turbines. Existing work has explored a range of supervised learning methods (\cite{mo2025review}). For example, \cite{kaya2019predicting} introduced a gas turbine PEMS dataset and compared three decision fusion schemes, highlighting the importance of certain features within the dataset for prediction, while \cite{LIU2020116627} used two surrogate models based on high dimensional model representation and artificial neural networks. \cite{10.1115/1.4065200} compared the performance of ensemble and conventional machine learning models. Gradient boosting methods such as XGBoost, AdaBoost, and CatBoost have been found to be successful in predicting gas turbine emissions (\cite{aslan2024prediction}, \cite{potts2023tabular}, \cite{DOSSANTOSCOELHO2024129366}).

However, most existing approaches evaluate predictive accuracy under labelled conditions, rather than whether individual predictions can be trusted when the model is deployed on unlabelled turbines. This distinction is important in fleet-level settings, where emissions labels may be available for only one or a small number of turbines, while the model is expected to generalise across units operating under different regimes. 

Self-supervised learning is attractive in this setting because it can use large quantities of unlabelled operational data to learn reusable representations, reducing reliance on labelled emissions measurements (\cite{zhao2024comparison}). However, industrial turbine data are noisy and collected across units operating under different operating conditions, such that the representations learned during self-supervised pre-training may capture useful shared dynamics, but may also reflect turbine-specific behaviour or domain shift. Therefore, self-supervised learning should not be assumed to improve downstream emissions prediction, and must be combined with uncertainty and trust diagnostics to assess whether predictions remain reliable on unlabelled turbines.

\subsection{Confidence Estimation and Prediction Reliability}

When predictive models are deployed under domain shift, point predictions alone are insufficient as they do not indicate whether a given prediction should be trusted. This is particularly important in fleet-level emissions monitoring, where models trained on labelled data from one turbine may be applied to unlabelled turbines operating under different regimes. Confidence estimation provides a way to assign a reliability measure to individual predictions, allowing uncertain or potentially unreliable outputs to be identified. 

Confidence has been used as a learned estimate of prediction reliability by \cite{devries2018learning}, in which network classifiers output confidence estimates for each input to differentiate between in and out-of-distribution samples, enabling the task to be completed without explicit labels. \cite{corbière2019addressingfailurepredictionlearning} introduced ConfidNet, which learns a confidence criterion based on the True Class Probability to identify predictions that are likely to fail, showing that learned confidence can provide a more meaningful failure indicator than relying directly on a model's raw prediction score. \cite{moon2020confidenceawarelearningdeepneural} framed confidence estimation in terms of ordinal ranking, where confidence values should distinguish correct predictions from likely errors. 

Early work by \cite{blatz2004confidence} demonstrated the use of confidence estimation for machine translation, where confidence scores were used to assess the likely correctness of word- and sentence-level outputs. \cite{lee2017training} proposed a confidence-calibrated training approach for out-of-distribution detection, encouraging classifiers to assign lower-confidence predictions to samples outside the training distribution.

\subsection{Uncertainty and Trust}

Uncertainty quantification provides an additional mechanism for identifying unreliable predictions. \cite{wang2025application} argue that uncertainty quantification can support out-of-distribution detection in safety-critical systems, helping identify cases where model predictions may be unreliable under unfamiliar operating conditions. 

Predictive uncertainty in machine learning is commonly divided into two main types: aleatoric uncertainty, arising from the inherent variability due to random effects; and epistemic uncertainty, reflecting uncertainty in the learnt model and its predictions (\cite{hullermeier2021aleatoric}).

Deep ensembles are a widely used approach for estimating predictive uncertainty. \cite{lakshminarayanan2017simple} showed that deep ensembles can improve uncertainty estimation and robustness under dataset shift, with out-of-distribution inputs generally receiving higher uncertainty. In addition to predictive uncertainty, feature-space distance has also been used for out-of-distribution detection. \cite{NEURIPS2018_abdeb6f5} proposed a Mahalanobis-distance-based confidence score derived from deep feature representations, supporting the use of latent-space distance as one component of the proposed trust framework. 

Trust-score approaches further motivate assessing reliability at the level of individual predictions, rather than evaluating only aggregate model accuracy. \cite{jiang2018trusttrustclassifier} define trust in terms of agreement with local neighbourhood structure, motivating the use of feature-space proximity and local consistency as indicators of prediction reliability. 

Together, these studies motivate the framework proposed in this work: a probabilistic emissions model that combines learned confidence, predictive uncertainty, feature-space distance, local consistency, and feature-prediction diagnostics to assess the reliability of individual predictions on unlabelled gas turbines.

\section{Materials and Methods}

This section defines the core challenge addressed in this work:  predicting emissions across a fleet of turbines using supervision from only a single labelled unit. However, turbines operate under different conditions and regimes, and therefore strong performance on the labelled turbine does not guarantee reliable predictions on unseen turbines. 

To address this, we aim to develop a framework that produces not only emissions predictions, but also calibrated trust scores for each prediction on unlabelled turbines. These trust scores are accompanied by interpretable, per-sample explanations, enabling assessment of prediction reliability in the absence of ground-truth emissions. This formulation reflects a practical deployment scenario in which emissions monitoring is unavailable or prohibitively expensive for the majority of turbines.

\subsection{Dataset}\label{sec:data}

The dataset consists of multivariate time series collected from 57 gas turbines. The labelled dataset is relatively small, comprising 107,936 sequences derived from a single gas turbine with available emissions measurements. In contrast, a substantially larger unlabelled dataset is available, containing 1,249,310 samples collected from turbines without emissions labels. Each sequence corresponds to a fixed-length input window of 64 time steps with a one-step prediction horizon.

Feature-level analysis highlights differences between labelled and unlabelled data distributions (Table \ref{tab:feature_stats}).

\begin{table}[H]
\centering
\caption{Normalised statistics of unlabelled data relative to the labelled training set.}
\begin{tabular}{lcc}
\hline
\textbf{Variable} & \textbf{Mean} & \textbf{Std} \\
\hline
TFire               & -0.50 & 1.92 \\
PT8                 & -1.00 & 2.07 \\
TC255               & -0.81 & 1.77 \\
TC2                 & -0.89 & 2.14 \\
GG\_Speed           &  0.14 & 1.37 \\
\hline
\label{tab:feature_stats}
\end{tabular}
\end{table}

\subsection{Prediction Model}\label{sec:pred_model}

To model gas turbine emissions and associated uncertainty, a multi-head probabilistic recurrent architecture based on an LSTM network is employed. The model jointly predicts the target emission, auxiliary feature dynamics, and confidence estimates, described in section \ref{sec:confidence}. 

Given an input sequence $x_{1:T} \in \mathbb{R}^{T \times d}$, where $T$ is the sequence length and $d$ is the number of input features, the sequence is processed using a stacked LSTM to predict the next step emissions target $y_{T+1}^{NOx}$.

The hidden representation from the final timestep, $h_T$, is used as a latent representation of the input sequence. Dropout is applied to this representation.

To account for turbine-specific behaviour, a learnable embedding is introduced. Each gas turbine index $g$ is mapped to a dense vector $e_g \in \mathbb{R}^k$, which is concatenated with the LSTM representation, allowing the model to capture systematic differences between turbines while sharing a common temporal representation. 

The emissions head predicts the mean and standard deviation for the target variable, while the feature head predicts the mean and standard deviation for each feature input at the next time step, $y_{T+1}^ {feat}$. This auxiliary feature task encourages the latent representation to capture the underlying system dynamics, particularly when labelled data are unavailable.

\subsection{Uncertainty Estimation}

An ensemble of $M$ independently initialised models is trained, where each model outputs a Gaussian predictive distribution for the target. Given an input sequence $x$, the $m$-th model produces a predictive mean $\mu_m(x)$ and variance $\sigma^2_m(x)$. At inference time, predictions are aggregated across the ensemble members, enabling the variability between models to be quantified.

\begin{equation}
    \mu(x)=\frac{1}{M}\sum^M_{m=1}\mu_m(x)
\end{equation}

Predictive uncertainty is decomposed into epistemic and aleatoric uncertainty. Epistemic uncertainty, representing uncertainty in the model parameters, is estimated as the variance of the ensemble predictions:

\begin{equation}
    \sigma^2_{ep}(x)=\frac{1}{M}\sum^M_{m=1}(\mu_m(x)-\mu(x))^2
\end{equation}

Aleatoric uncertainty, corresponding to the inherent data noise, is computed as the mean of the predictive variances output by individual models:

\begin{equation}
    \sigma^2_{al}(x)=\frac{1}{M}\sum^M_{m=1}\sigma^2_m(x)
\end{equation}

The total predictive uncertainty is then defined as the sum of these components:

\begin{equation}
    \sigma^2_{total}(x)=\sigma^2_{ep}(x)+\sigma^2_{al}(x)
\end{equation}

This decomposition provides a principled separation between uncertainty arising from limited model knowledge and irreducible data noise.

\subsection{Confidence Estimation}\label{sec:confidence}

Inspired by learned confidence estimation approaches (\cite{devries2018learning}), an additional confidence head is trained jointly with the emissions prediction task. For each input sequence, the model outputs: an emissions mean prediction, $\hat{y}$; an emissions uncertainty estimate, $\hat{\sigma}$; and a scalar confidence value, $c \in [0,1]$. The emissions head is trained probabilistically using Gaussian negative log likelihood loss, encouraging the model to predict both accurate means and appropriate predictive variance:

\begin{equation}
    \mathcal{L}_{em}=\mathcal{L}_{GNLL}(\hat{y}, y, \hat{\sigma^2})
    \label{eq:GNLL}
\end{equation}

where $y$ is the ground-truth emissions target. 

To train the confidence head, the prediction error is first converted back into the original emissions unit. Let $\hat{y}_{orig}$ and $y_{orig}$ denote the inverse-transform prediction and target. The absolute error is then computed as

\begin{equation}
    e=|\hat{y}_{orig}-y_{orig}|
    \label{eq:absolute_error}
\end{equation}

A soft confidence target is constructed from the prediction error:

\begin{equation}
    c^* = exp(-\frac{e}{\tau})
    \label{eq:confidence_target}
\end{equation}

Where $\tau$ is a scaling parameter in the original emissions units, such that errors small relative to this tolerance receive high confidence targets. We have set $\tau=5$. This formulation assigns confidence values close to 1 for low-error predictions and smaller values as the error increases. 

The predicted confidence $c$ is constrained to $[0,1]$ through a sigmoid operation, and is then trained against this target using a binary cross-entropy loss, such that confidence reflects the likelihood of low prediction error:

\begin{equation}
    \mathcal{L}_{conf}=BCE(c,c^*).
    \label{eq:bce_loss}
\end{equation}

In addition to the emissions confidence, a feature-level confidence head is trained to estimate the reliability of the auxiliary feature predictions. For each feature $x_i$, a confidence target is defined analogously:

\begin{equation}
    c^*_{x_i} = \exp\left(-\frac{|x_i - \hat{x}_i|}{\tau_i}\right)
\end{equation}

where $\tau_i$ is a feature-specific scaling factor derived from the training distribution. The feature confidence outputs are trained using binary cross-entropy in the same manner as the emissions confidence.

The overall training objective combines emissions prediction, feature prediction, and confidence supervision.

\begin{equation}
\mathcal{L} =
\mathcal{L}_{em}
+ \lambda_{feat} \mathcal{L}_{feat}
+ \lambda_{conf} \mathcal{L}_{conf}
+ \lambda_{conf\_feat} \mathcal{L}_{conf\_feat}
\end{equation}

where $\mathcal{L}_{feat}$ is the Gaussian negative log-likelihood for feature prediction, and $\mathcal{L}_{conf\_feat}$ is the feature-level confidence loss, where the weighting terms control the relative contribution of each auxiliary objective, ensuring that emissions prediction remains the primary task.

\subsection{Trust Score}\label{sec:trust_score}

The trust score is derived from an error-calibrated model trained on the labelled validation set. For each validation sample, four reliability signals are extracted: Mahalanobis distance, $D^2$, predictive uncertainty, $\sigma$, epistemic uncertainty, and emissions confidence. These signals are used as inputs to an XGBoost regression model trained to predict the observed absolute prediction error. 

The trained error model produces a combined expected error for each sample which represents an estimate of prediction risk based on the joint behaviour of the uncertainty and confidence signals. Trust thresholds are then calibrated from the distribution of predicted errors on the labelled validation set. 

The combined expected error is mapped to a continuous trust score in the range $[0,100]$. This score is subsequently used to assign discrete trust levels, enabling both the fine-grained and categorical assessment of prediction reliability, providing an interpretable value for prediction quality. 

Input features are also compared against the labelled training operating range. A feature is considered outside the range when more than 5\% of the time steps in the sequence fall outside the observed training bounds. The number of out-of-range features is reported as an additional diagnostic signal and included in textual explanations of reduced trust.

\subsection{Training Procedure}\label{sec:training}

To leverage both labelled and unlabelled data, a two-stage training procedure is adopted. 

\textbf{Stage 1: Unlabelled pretraining.}
The model is first trained using only the feature prediction and feature confidence objectives. This stage encourages the model to learn general system dynamics without requiring emissions labels.

\textbf{Stage 2: Labelled training.}
The model is then trained on labelled data using the full loss function. Emissions prediction, feature prediction, and both confidence heads are optimised jointly. 

This two-stage approach allows the model to benefit from large volumes of unlabelled data while anchoring confidence calibration to the labelled dataset, enabling confidence estimates to act as a proxy for prediction error when ground-truth labels are unavailable.

\subsection{Trust Validation}\label{sec:trust_eval}

Building on the uncertainty and reliability signals defined above, overall trustworthiness is assessed at both model and domain levels. To assess confidence in predictions on unlabelled gas turbines, both the model's prediction performance on labelled data and its ability to bridge potential domain gaps must be validated. This is achieved through rigorous evaluation using both labelled and unlabelled datasets.

\subsubsection{Model Reliability}\label{sec:model_trust}

Model-level trustworthiness is first assessed using standard validation techniques, including hold-out validation to compute mean absolute error (MAE) and root mean squared error (RMSE), alongside visual inspection of predicted versus true emissions. Same-model ensembling is used to evaluate the consistency of prediction under different random initialisations.

More advanced approaches are employed to explicitly quantify model uncertainty. Prediction interval coverage probability (PICP) is used to measure the proportion of true targets that fall within a nominal 95\% prediction interval, providing insight into the reliability of uncertainty estimates. This is analysed in conjunction with the mean prediction interval width (MPIW) to assess whether prediction intervals are appropriately narrow across different operating regimes. To avoid trivial solutions with overly wide intervals, the uncertainty ratio is used to penalise excessive interval widths while maintaining target coverage. Meaningful uncertainty estimates are indicated by low-uncertainty predictions corresponding to high accuracy and high-uncertainty predictions corresponding to lower accuracy. 

To define the uncertainty ratio, let the prediction be $\hat{y}_i$, predicted standard deviation $\sigma_i$, and the true value $y_i$. The prediction interval is therefore

\begin{equation}
    Lower_i=\hat{y}_i-z\cdot\sigma_i
\end{equation}

and

\begin{equation}
    Upper_i=\hat{y}_i+z\cdot\sigma_i
\end{equation}

Thus the MPIW is given as: 

\begin{equation}
    MPIW=\frac{1}{N}\sum^N_{i=1}(Upper_i-Lower_i)
    \label{eq:MPIW}
\end{equation}

The PICP is given as 

\begin{equation}
    PICP=\frac{1}{N}\sum^N_{i=1}c_i
    \label{eq:PICP}
\end{equation}

where $c_i$ is a binary indicator variable for sample $i$ to determine if a sample is between $Lower_i$ and $Upper_i$. 

The uncertainty ratio is therefore given as 

\begin{equation}
    \mathrm{Uncertainty\ Ratio} = \frac{MPIW}{2\cdot1.96\cdot\sigma_y}
\end{equation}

where $\sigma_y=\mathrm{std}(y)$.

These metrics allow us to ensure the model itself has learned appropriate predictions on both labelled and unlabelled data.

\subsubsection{Domain Reliability}\label{sec:domain_trust}

To assess the trustworthiness of predictions on unlabelled data, feature space analysis and comparative evaluation between labelled and unlabelled predictions are performed. 

Out-of-distribution (OOD) analysis is used to assess extrapolation reliability. Mahalanobis distance ($D^2$) is computed between each sample's latent representation and the mean distribution of training sequences in the model's feature space. A 99\% $D^2$ threshold on the labelled training data is used to classify samples as in-distribution (ID) or out-of-distribution (OOD).

Proxy-based validation is also employed by training the model to jointly predict emissions and future measurable features, incorporating both labelled and unlabelled data into the loss function. During inference on both labelled and unlabelled data, similar to predicting emissions, next step feature data is predicted. The accuracy of feature prediction serves as a proxy for confidence in the associated emissions prediction. 

Operational constraint-based validation is applied through range analysis to ensure that predictions are made within physically plausible operating regimes. Input features are compared against the observed training bounds, and deviations are used as an indicator of extrapolation beyond known system behaviour.

\subsection{PEMS Diagnostic Report}

To provide interpretable and actionable outputs, all reliability signals are consolidated into a per-sample diagnostic report. For each prediction, the report includes the trust score and the underlying signals used to derive this assessment. 

In addition to numerical values, the report provides concise textual explanations highlighting the primary drivers of reduced trust, such as elevated uncertainty or model disagreement. These explanations are derived directly from underlying signals and are designed to support rapid interpretation without requiring inspection of individual metrics. 

This formulation enables the trust framework to function as a diagnostic tool, allowing users to assess prediction reliability as well as understand the contributing factors behind each assessment, essential for deployment in real-world emissions monitoring systems.

\section{Results}

\begin{figure}[h!]
\centering
\begin{subfigure}[b]{0.48\textwidth}
    \centering
    \includegraphics[width=\linewidth]{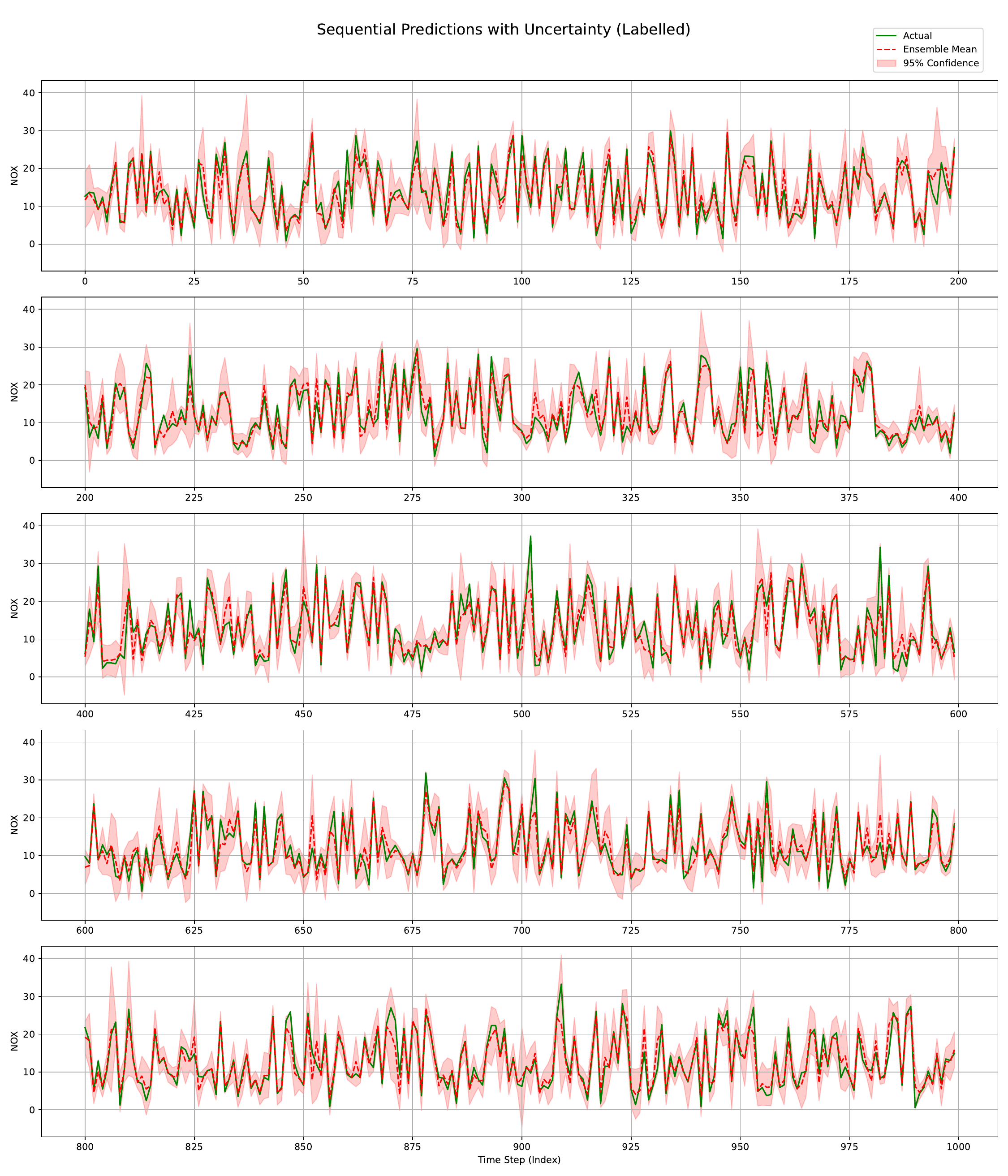}
    \caption{Labelled data.}
    \label{fig:sample_predictions_labelled}
\end{subfigure}
\hfill
\begin{subfigure}[b]{0.48\textwidth}
    \centering
    \includegraphics[width=\linewidth]{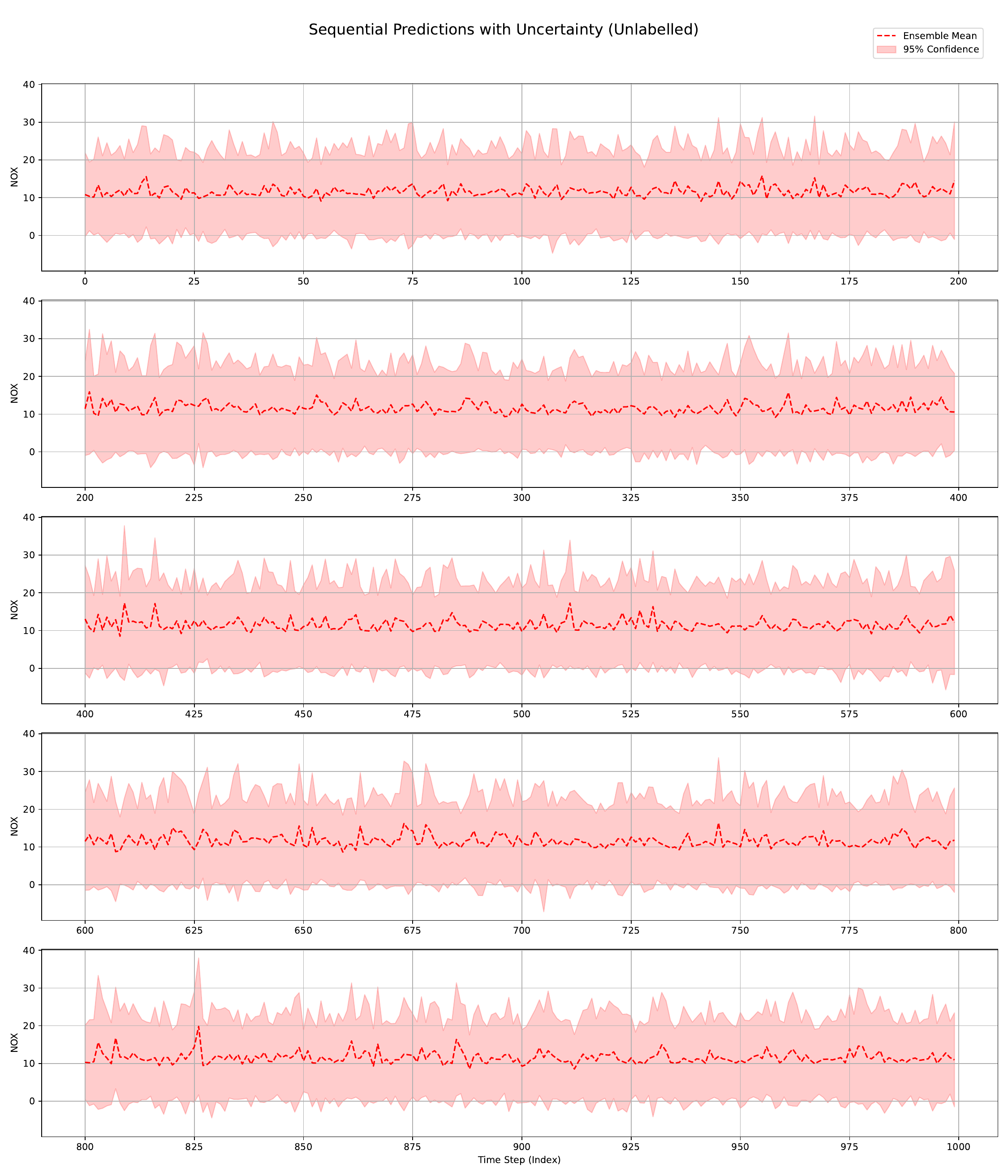}
    \caption{Unlabelled data.}
    \label{fig:sample_predictions_unlabelled}
\end{subfigure}
\caption{Sequential NOx predictions with 95\% prediction intervals for \textbf{(A)} labelled data and \textbf{(B)} unlabelled data.}
\label{fig:sample_predictions}
\end{figure}

\subsection{Model Performance and Uncertainty Behaviour}\label{sec:model_results}

A key requirement of a reliable model is that its confidence estimates meaningfully reflect prediction error. On the labelled dataset, an average MAE of 0.202 and RMSE of 0.303 were obtained. This performance is comparable to an XGBoost baseline (MAE 0.313, RMSE 0.435), demonstrating that the probabilistic model maintains competitive predictive accuracy. Example predictions are shown in Figure \ref{fig:sample_predictions}, where the predicted emissions closely reflect the ground-truth emissions with narrow uncertainty bands. In contrast, the unlabelled dataset exhibits substantially higher uncertainty, with much larger prediction intervals. This increase reflects the model's reduced certainty when applied to unseen operating conditions, highlighting the need for additional reliability signals when deploying predictions under distributional shift.

To further assess whether uncertainty behaves appropriately, sequences were analysed for in-distribution (ID) and out-of-distribution (OOD) samples, where OOD is defined using the 99th percentile Mahalanobis distance threshold of the labelled training data. As shown in Table \ref{tab:MPIW}, the normalised mean prediction width (NMPIW) increases consistently from labelled to unlabelled data and from ID to OOD samples, indicating that the model appropriately expands its uncertainty under distributional shift.

\begin{table}[H]
\centering
\caption{Normalised mean rediction interval width for in-distribution and out-of-distribution sequences.}
\begin{tabular}{lcc}
\hline
\textbf{Dataset} & \textbf{NMPIW (ID)} & \textbf{NMPIW (OOD)} \\
\hline
Labelled   & 1.01 & 2.04 \\
Unlabelled & 7.88 & 10.75 \\
\hline
\label{tab:MPIW}
\end{tabular}
\end{table}

Overall, these results demonstrate that the model achieves accurate predictions on labelled data while responding appropriately to unfamiliar operating conditions through increased predictive uncertainty, providing a suitable foundation for subsequent trust evaluation.

\subsection{Confidence Calibration}\label{sec:confidence_results}

\begin{figure}[h!]
\begin{center}
\includegraphics[width=10cm]{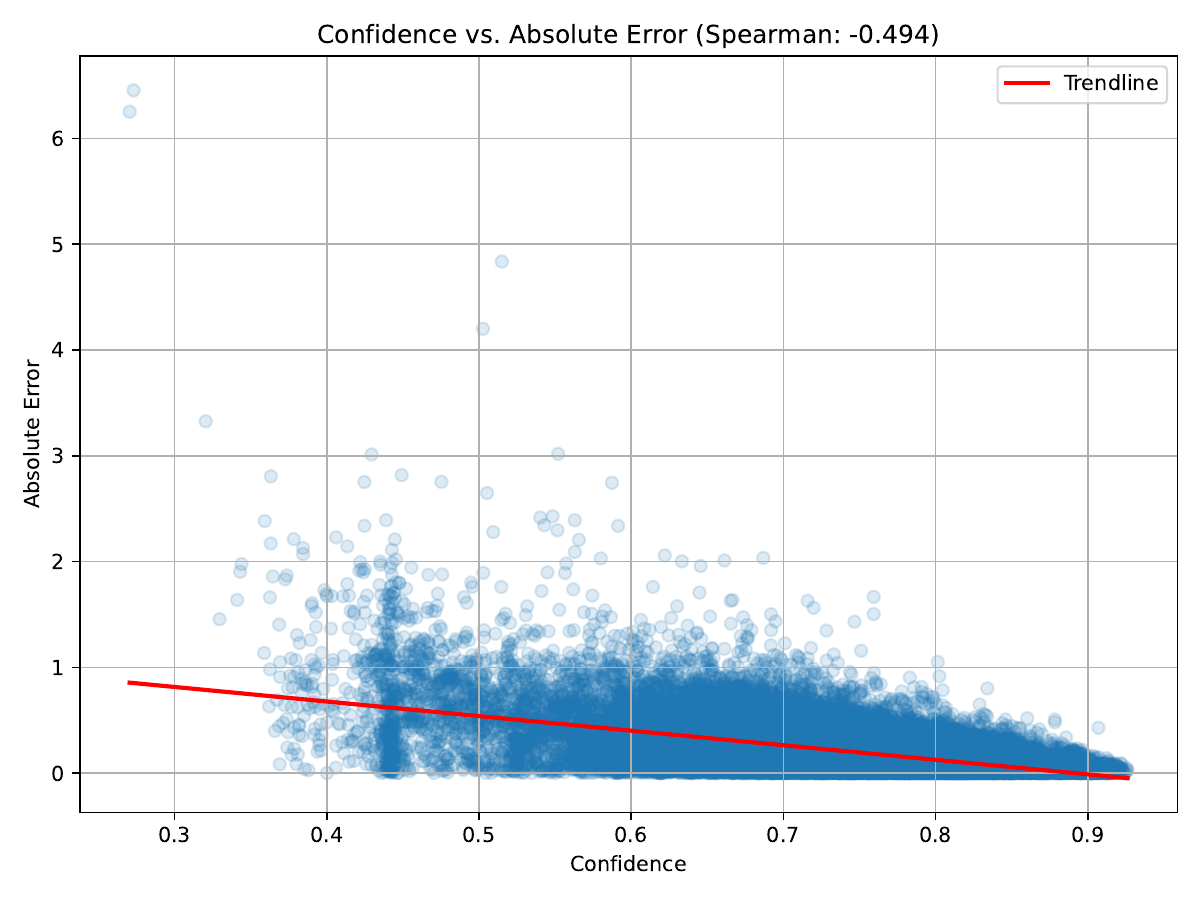}
\end{center}
\caption{Predicted confidence compared to normalised prediction error on labelled data.}\label{fig:confidence_vs_error}
\end{figure}

Confidence is strongly aligned with prediction error. Figure \ref{fig:confidence_vs_error} shows a clear inverse relationship between confidence and error on the labelled dataset, demonstrating that the model's confidence estimates are meaningful. This implies that high-confidence predictions can be trusted to have low error, which is essential for deployment in real-world settings.

\begin{figure}[h!]
\begin{center}
\includegraphics[width=10cm]{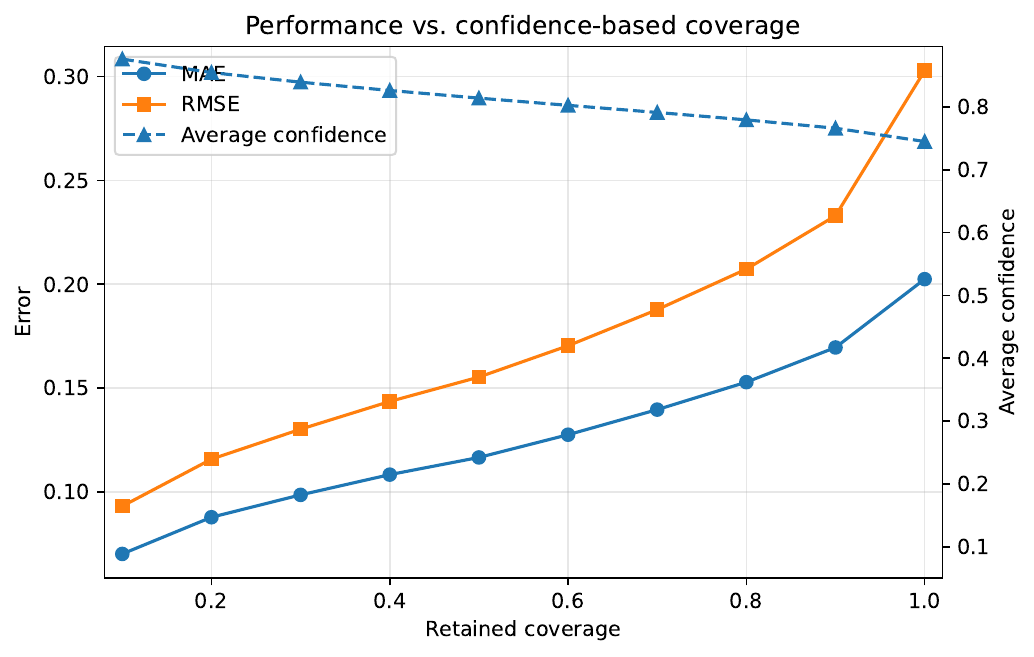}
\end{center}
\caption{Prediction error (MAE and RMSE) as a function of retained high-confidence samples. Performance improves consistently as lower-confidence predictions are removed.}\label{fig:performance_vs_confidence_coverage}
\end{figure}

This behaviour is further validated through confidence-based filtering. Table \ref{tab:performance_vs_confidence_coverage} shows that restricting evaluation to higher-confidence predictions leads to consistent improvements in performance. As coverage decreases from 100\% to 10\%, MAE reduces from 0.2024 to 0.0701, demonstrating that confidence provides a reliable mechanism for identifying accurate predictions. This trend is further illustrated in Figure \ref{fig:performance_vs_confidence_coverage}. This confirms that confidence is not only correlated with error, but is also practically useful for decision-making.

\begin{table}[H]
\centering
\caption{Performance compared to confidence-based coverage. Metrics are evaluated on the top-$k$ highest-confidence predictions, showing improved accuracy as low-confidence samples are removed.}
\begin{tabular}{lcccc}
\hline
\textbf{Coverage} & \textbf{MAE} & \textbf{RMSE} & \textbf{Confidence} \\
\hline
1.0 & 0.2024 & 0.3029 & 0.7449 \\
0.9 & 0.1695 & 0.2330 & 0.7660 \\
0.8 & 0.1528 & 0.2073 & 0.7793 \\
0.7 & 0.1396 & 0.1877 & 0.7911 \\
0.6 & 0.1275 & 0.1704 & 0.8025 \\
0.5 & 0.1166 & 0.1553 & 0.8139 \\
0.4 & 0.1083 & 0.1435 & 0.8261 \\
0.3 & 0.0985 & 0.1302 & 0.8393 \\
0.2 & 0.0878 & 0.1157 & 0.8546 \\
0.1 & 0.0701 & 0.0931 & 0.8761 \\
\hline
\label{tab:performance_vs_confidence_coverage}
\end{tabular}
\end{table}

\begin{figure}[h!]
\centering
\begin{subfigure}[b]{0.48\textwidth}
    \centering
    \includegraphics[width=\linewidth]{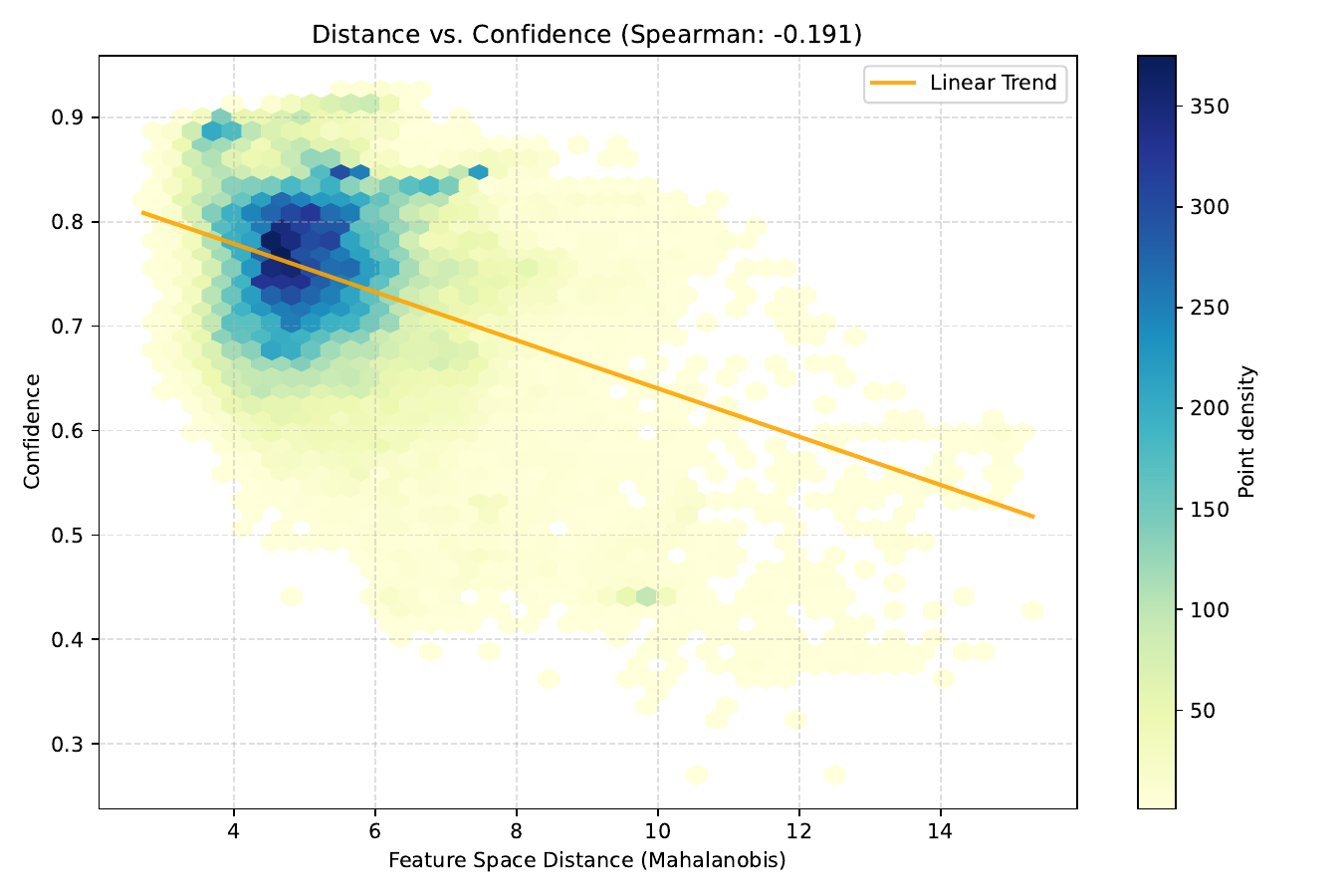}
    \caption{Labelled data.}
    \label{fig:dist_vs_conf_labelled}
\end{subfigure}
\hfill
\begin{subfigure}[b]{0.48\textwidth}
    \centering
    \includegraphics[width=\linewidth]{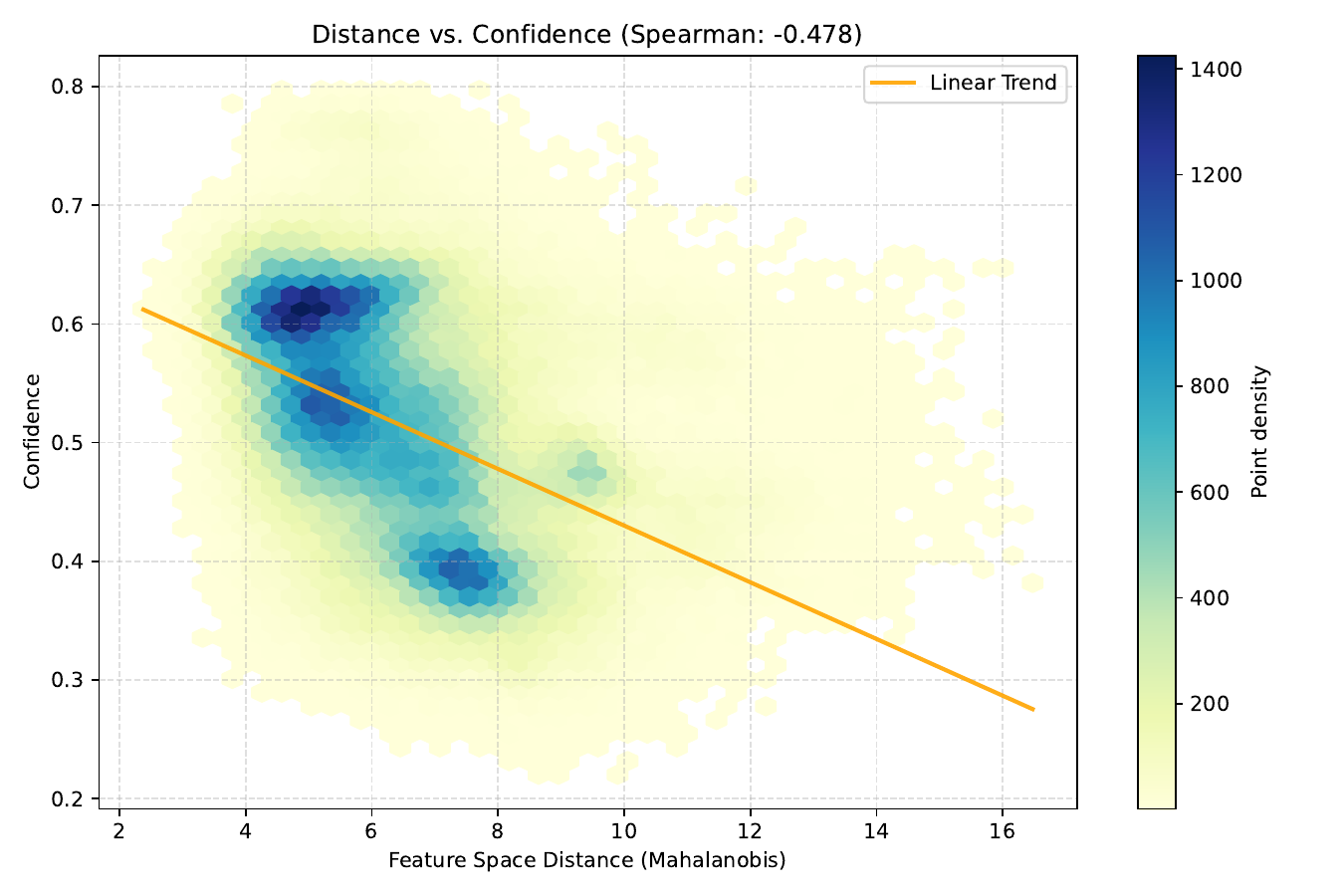}
    \caption{Unlabelled data.}
    \label{fig:dist_vs_conf_unlabelled}
\end{subfigure}
\caption{Relationship between Mahalanobis distance and predicted emissions confidence for \textbf{(A)} labelled data and \textbf{(B)} unlabelled data.}
\label{fig:distance_vs_confidence}
\end{figure}

This relationship can be further analysed by examining how confidence varies with distance from the training distribution. As shown in Table \ref{tab:confidence_vs_distance_comparison}, confidence generally decreases as Mahalanobis distance increases for both labelled and unlabelled data. This trend is more pronounced in the labelled dataset, where confidence drops as distance increases. For the unlabelled data, a similar but weaker trend is observed, with confidence decreasing across most distance ranges. However, at the highest distances, confidence becomes less consistent, likely due to the small number of samples in these bins. This suggests that while the model is sensitive to distributional shift, confidence alone does not fully capture extreme OOD behaviour.
Figure \ref{fig:distance_vs_confidence} shows that confidence decreases as distance from the training distribution increases. This trend is more consistent in the labelled dataset, while the unlabelled dataset exhibits greater variability, particularly at high distances where fewer samples are available. 
The overall negative relationship between confidence and distance indicates that the model appropriately reduces confidence as samples move further from the training distribution, supporting the inclusion of both signals within the trust framework.

\begin{table}[H]
\centering
\caption{Confidence compared to distance comparison (labelled vs. unlabelled).}
\begin{tabular}{lccc}
\hline
\textbf{Distance Range} & \textbf{Num Samples (L / U)} & \textbf{Mean Conf. (L)} & \textbf{Mean Conf. (U)} \\
\hline
2.4-3.8   & 2892 / 4760   & 0.79 & 0.59 \\
3.8-5.2   & 19540 / 59361 & 0.76 & 0.57 \\
5.2-6.6   & 15269 / 79417 & 0.75 & 0.54 \\
6.6-8.0   & 5306 / 63827  & 0.72 & 0.47 \\
8.0-9.4   & 1242 / 27456  & 0.68 & 0.45 \\
9.4-10.8  & 722 / 9912    & 0.56 & 0.46 \\
10.8-12.3 & 170 / 3801    & 0.52 & 0.47 \\
12.3-13.7 & 104 / 985     & 0.49 & 0.47 \\
13.7-15.1 & 66 / 252      & 0.55 & 0.48 \\
15.1-16.5 & 1 / 51        & 0.42 & 0.47 \\
\hline
\label{tab:confidence_vs_distance_comparison}
\end{tabular}
\end{table}

\begin{figure}[h!]
\centering
\begin{subfigure}[b]{0.48\textwidth}
    \centering
    \includegraphics[width=\linewidth]{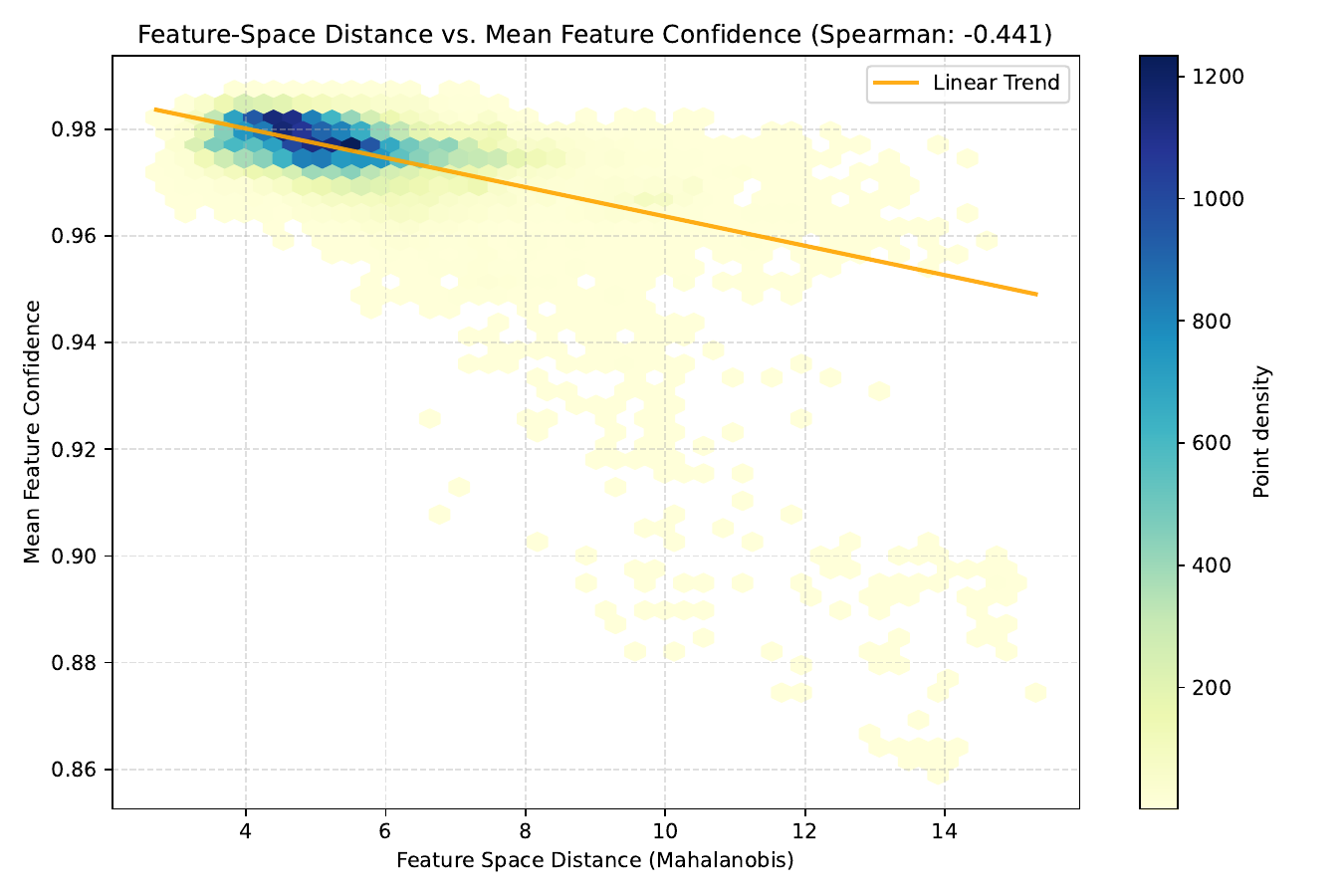}
    \caption{Labelled data.}
    \label{fig:distance_vs_feature_confidence_labelled}
\end{subfigure}
\hfill
\begin{subfigure}[b]{0.48\textwidth}
    \centering
    \includegraphics[width=\linewidth]{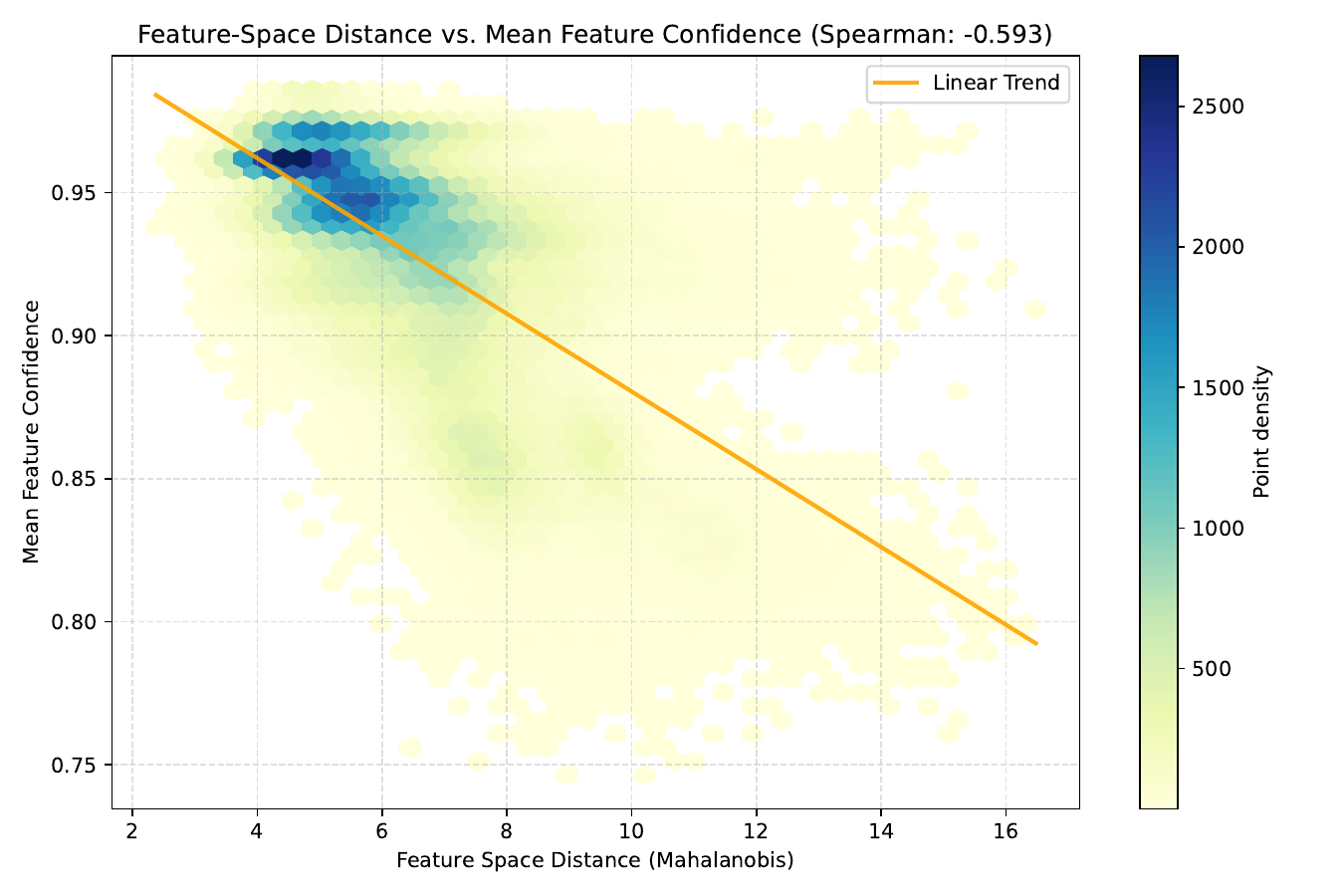}
    \caption{Unlabelled data.}
    \label{fig:distance_vs_feature_confidence_unlabelled}
\end{subfigure}
\caption{Relationship between Mahalanobis distance and mean feature confidence for \textbf{(A)} labelled data and \textbf{(B)} unlabelled data.}
\label{fig:distance_vs_feature_confidence}
\end{figure}

In addition to emissions confidence, we further compare the average predicted feature confidence to the distance. Figure \ref{fig:distance_vs_feature_confidence} shows trends similar to those observed for emissions confidence, showing that the feature confidence predictions can also be used as a signal for the trust framework.

Importantly, these results demonstrate that trustworthiness cannot be captured using a single signal alone. Mahalanobis distance provides information about distributional unfamiliarity, while confidence estimates prediction reliability. These signals provide complementary information required for robust trust assessment under distributional shift.

\subsection{Trust Framework Evaluation}\label{sec:trust_framework_evaluation}

Table \ref{tab:pems_report} presents an example diagnostic output generated by the proposed trust framework for a medium-trust prediction. In addition to the overall trust score and trust level, the report provides the underlying reliability signals used to derive the assessment. To improve interpretability, concise textual explanations are generated directly from the reliability signals, highlighting the primary factors contributing to reduced trust. This enables users to understand not only the overall trust assessment, but also the underlying causes of uncertainty or reduced reliability.

\begin{table}[H]
\centering
\caption{Example trust report for a medium-trust prediction.}
\label{tab:pems_report}
\begin{tabular}{ll}
\toprule
\textbf{Component} & \textbf{Value} \\
\midrule
Trust Score & 65 \\
Trust Level & Medium \\
\midrule
Confidence & 0.615 \\
Predictive Uncertainty ($\sigma$) & 0.571 \\
Epistemic Uncertainty & 0.080 \\
Mahalanobis Distance ($D^2$) & 5.177 \\
Feature Confidence (mean) & 0.974 \\
Feature Confidence (min) & 0.884 \\
Features Outside Range & 0 \\
\midrule
Reasons & Moderate model disagreement \\
        & Moderate predictive uncertainty \\
        & Low emissions confidence \\
\bottomrule
\end{tabular}
\end{table}

Table \ref{tab:trust_comparison} evaluates the proposed trust framework across labelled and unlabelled datasets. There is a small number of low-trust samples in the labelled set as the trust thresholds are calibrated using operating conditions already observed during training. On the labelled data, a clear relationship is observed between trust level and prediction error, with mean error increasing as trust decreases. This demonstrates that the trust score provides a meaningful ranking of prediction reliability on samples where ground-truth emissions are available.

\begin{table}[H]
\centering
\caption{Trust-level comparison for labelled and unlabelled datasets. Metrics are averaged per trust level, demonstrating the relationship between trust score, uncertainty signals, and prediction error.}
\begin{tabular}{lccccccccc}
\toprule
\textbf{Level} & \textbf{N} & \textbf{Trust} & \textbf{Confidence} & $\boldsymbol{\sigma}$ & \textbf{Epistemic} & $\mathbf{D^2}$ & \makecell{\textbf{Feat Pred}\\\textbf{Error}} & \makecell{\textbf{Feats Outside}\\\textbf{Range}} & \textbf{Err} \\

\midrule

\multicolumn{10}{l}{\textbf{Labelled}} \\
High & 6401 & 93.02 & 0.74 & 0.32 & 0.01 & 5.49 & 0.15 & 0.08 & 0.22 \\
Med  & 68   & 65.46 & 0.48 & 0.92 & 0.15 & 7.50 & 0.46 & 0.29 & 0.77 \\
Low  & 4    & 45.47 & 0.40 & 1.22 & 0.28 & 8.88 & 0.17 & 0.00 & 1.11 \\

\midrule

\multicolumn{10}{l}{\textbf{Unlabelled}} \\
High & 111652 & 81.65 & 0.58 & 0.66 & 0.10 & 6.07 & 0.66 & 0.98 & -- \\
Med  & 128825 & 64.36 & 0.46 & 0.95 & 0.15 & 6.87 & 1.34 & 2.80 & -- \\
Low  & 9346   & 42.29 & 0.42 & 1.05 & 0.12 & 5.95 & 1.55 & 2.57 & -- \\

\bottomrule
\label{tab:trust_comparison}
\end{tabular}
\end{table}

Across both datasets, lower trust levels are associated with reduced confidence, increased predictive uncertainty, and higher epistemic uncertainty compared to high-trust samples. Similarly, the number of features operating outside the labelled training range increases for lower-trust predictions. The consistent behaviour of these independent signals suggests that the framework captures multiple complementary aspects of prediction reliability rather than relying on a single uncertainty estimate. This is particularly evident in the unlabelled dataset, where the average Mahalanobis distance alone does not monotonically increase with decreasing trust levels, indicating that the distributional distance by itself is insufficient to fully characterise prediction reliability. 

The coherence between uncertainty, confidence, feature behaviour, and operating-range diagnostics suggests that the framework provides a consistent reliability assessment under distributional shift, even in the absence of emissions labels on the unlabelled turbines.

\begin{figure}[h!]
\begin{center}
\includegraphics[width=10cm]{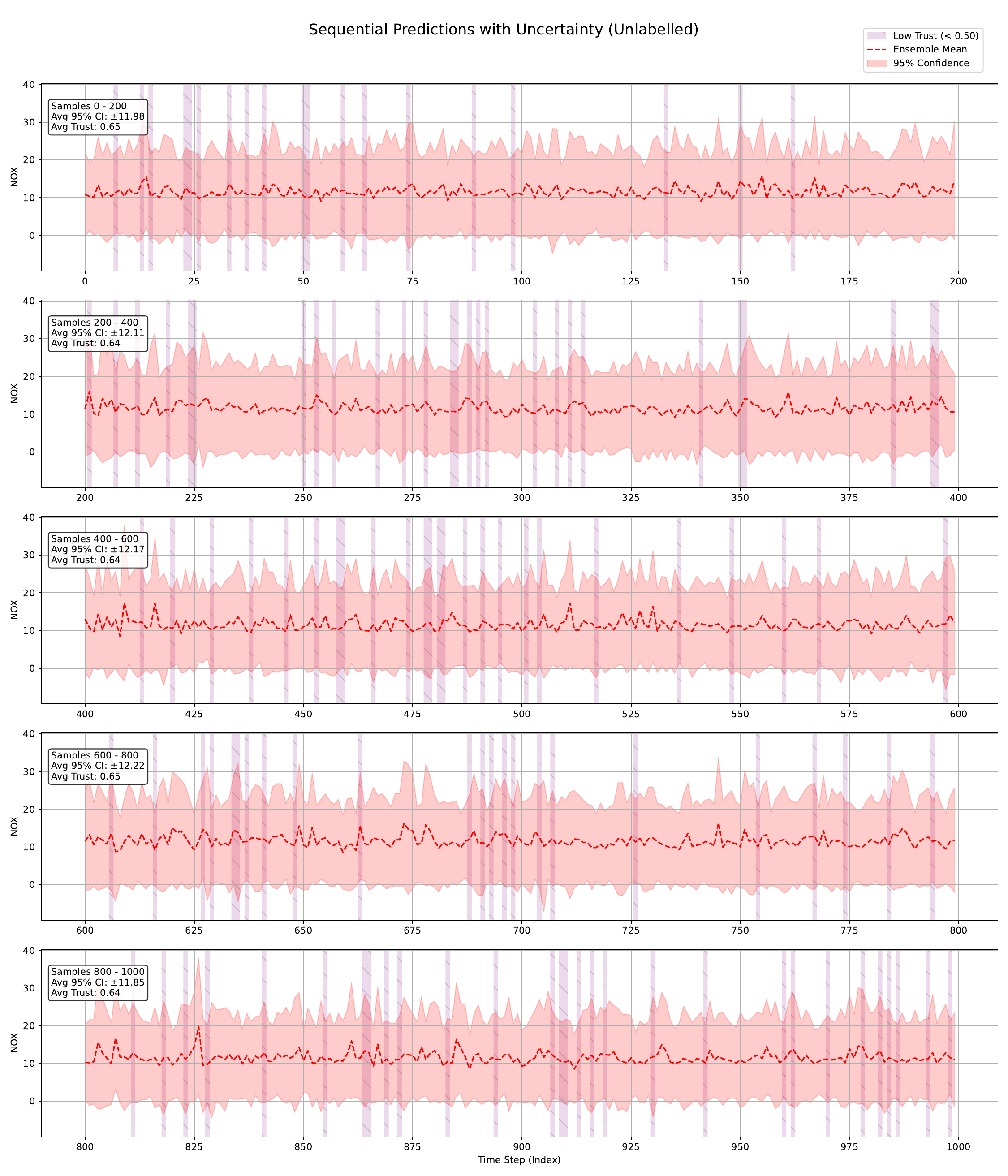}
\end{center}
\caption{Sequential predictions for unlabelled data with low-trust regions highlighted.}\label{fig:preds_trust_unlabelled}
\end{figure}

Figure \ref{fig:preds_trust_unlabelled} illustrates the regions in which low trust scores are assigned within the sequential predictions. This provides an interpretable visual indication of where the model considers predictions to be less reliable within the sequence, enabling potentially problematic operating regions to be identified.

\section{Discussion}

The results demonstrate that the proposed framework can support fleet-level gas turbine emissions prediction when emissions labels are available for only a limited subset of assets. While the probabilistic LSTM achieved strong performance on the labelled data, the main contribution of this work is the addition of prediction-level trust assessment. This is important because accurate performance on a labelled test set does not guarantee reliable predictions on unlabelled turbines operating under different conditions.

Most existing machine learning approaches for predictive emissions monitoring focus on aggregate prediction accuracy under labelled evaluation conditions. In contrast, this work addresses whether individual predictions can be trusted when ground-truth emissions measurements are unavailable. By combining learned confidence, predictive uncertainty, epistemic uncertainty, feature-space distance, auxiliary feature prediction, and operating-range diagnostics, the framework provides a broader assessment of reliability than accuracy metrics alone. The confidence-based results show that learned confidence is meaningfully related to prediction error. As lower-confidence samples are removed, MAE and RMSE decrease, indicating that confidence can identify more reliable predictions. However, confidence alone is not sufficient under distributional shift. The unlabelled data show increased uncertainty and reduced confidence, while out-of-distribution samples have wider prediction intervals. This suggests that the model responds appropriately to unfamiliar operating conditions by expressing greater uncertainty.

The trust-level analysis further supports the value of combining multiple reliability signals. On labelled data, lower trust levels correspond to higher prediction error, validating the trust score where ground truth is available. On unlabelled data, true emissions error cannot be measured, but lower trust levels are associated with reduced confidence, increased uncertainty, higher feature prediction error, and more frequent deviations from the labelled operating range. This provides an interpretable indication of where predictions should be treated with caution. A key implication is that trustworthiness in fleet-level PEMS should be treated as a multi-signal problem. Mahalanobis distance indicates whether a sample is far from the labelled training distribution, but does not directly measure predictive reliability. Confidence, uncertainty, feature prediction error, and operating-range diagnostics each provide complementary information. Their combination therefore gives a more robust basis for reliability assessment than any single signal alone.

There are several limitations to this study. First, emissions labels are available from only one turbine, so prediction error cannot be directly validated on the unlabelled turbines. The trust score should therefore be interpreted as a calibrated estimate of reliability rather than a direct measurement of accuracy. Second, the trust thresholds are derived from the labelled validation data and depend on the representativeness of this labelled turbine. Finally, although the framework is intended to be model-agnostic, it has only been evaluated here using a probabilistic LSTM backbone.

Overall, this work shows that trustworthy fleet-level emissions prediction requires more than accurate point estimates. The proposed framework provides interpretable, per-sample trust assessments that indicate when predictions on unlabelled turbines are likely to be reliable and when they should be treated with caution.

\section{Conclusion}\label{sec:conclusion}

This work proposed a trust-aware probabilistic PEMS framework for fleet-level gas turbine emissions prediction where emissions labels are available for only one turbine. The framework combines learned confidence, ensemble uncertainty, feature-space distance, feature-prediction diagnostics, and operating-range analysis to produce calibrated per-sample trust scores. Results on labelled data showed strong predictive performance and demonstrated that confidence and trust scores were meaningfully associated with prediction error. On unlabelled turbines, increased uncertainty and reduced confidence indicated that the framework responds to distributional shift and provides interpretable reliability information when ground-truth emissions are unavailable. Future work should validate the framework on additional labelled turbines when such data becomes available. In addition, further work should explore stronger backbone models to improve predictive performance, since the proposed trust framework is designed to be model-agnostic and transferable across predictive architectures.

\section*{Conflict of Interest Statement}

The authors declare that the research was conducted in the absence of any commercial or financial relationships that could be construed as a potential conflict of interest.

\section*{Funding}

The work presented here received funding from EPSRC (EP/W522089/1) and Siemens Energy Industrial Turbomachinery Ltd. as part of the iCASE EPSRC PhD studentship “Predictive Emission Monitoring Systems for Gas Turbines”.

\bibliographystyle{Frontiers-Harvard}
\bibliography{references}

\end{document}